# IEEE Copyright Notice

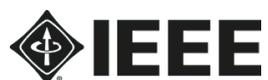

# Incremental Multistep Forecasting of Battery Degradation Using Pseudo Targets

Jonathan Adam Rico[§]
Engineering Product Development
SUTD, Singapore
Institute for Infocomm Research
(I²R), A*STAR, Singapore
jonathanadam_rico@mymail.sutd.edu.sg
[§]corresponding author

Nagarajan Raghavan
Engineering Product Development
SUTD, Singapore
nagarajan@sutd.edu.sg

Senthilnath Jayavelu
Institute for Infocomm Research (I²R)
A*STAR, Singapore
j_senthilnath@i2r.a-star.edu.sg

*Abstract*—Data-driven models accurately perform early battery prognosis to prevent equipment failure and further safety hazards. Most existing machine learning (ML) models work in offline mode which must consider their retraining post-deployment every time new data distribution is encountered. Hence, there is a need for an online ML approach where the model can adapt to varying distributions. However, existing online incremental multistep forecasts are a great challenge as there is no way to correct the model of its forecasts at the current instance. Also, these methods need to wait for a considerable amount of time to acquire enough streaming data before retraining. In this study, we propose iFSNet (incremental Fast and Slow learning Network) which is a modified version of FSNet for a single-pass mode (sample-by-sample) to achieve multistep forecasting using pseudo targets. It uses a simple linear regressor of the input sequence to extrapolate pseudo future samples (pseudo targets) and calculate the loss from the rest of the forecast and keep updating the model. The model benefits from the associative memory and adaptive structure mechanisms of FSNet, at the same time the model incrementally improves by using pseudo targets. The proposed model achieved 0.00197 RMSE and 0.00154 MAE on datasets with smooth degradation trajectories while it achieved 0.01588 RMSE and 0.01234 MAE on datasets having irregular degradation trajectories with capacity regeneration spikes.

## I. INTRODUCTION

Lithium-ion batteries (LIBs) power our electronic devices and electric vehicles over the past decades. However, LIBs degrade over time and may need to be replaced before it is too late for its second life applications [1]. Thus, a battery management system (BMS) is necessary to perform many tasks including prognosis for early prevention of device or equipment failure and other safety risks [2]. Data-driven models became popular due to the improved speed and accuracy in battery prognosis as compared to empirical and model-based methods [3]. Traditionally, a data-driven model is trained on collected historical data and then deployed to BMS. However, data distribution varies over time from the trained historical data which necessitates retraining. Later, online learning began to emerge [4] so that models continue to improve post-deployment by training on new data in an online scheme. For example, deployed chatbots and video recommendation ML models perform poorly at the beginning but start to improve as people continuously input data into these models.

In online learning, data comes in a stream as a single-pass mode (sample-by-sample), and thus only the next-step ahead forecast can be trained, the rest of the multistep ahead forecast cannot be optimized. Stefani *et al.* [5] developed a dynamic factor machine learner algorithm for incremental multistep ahead forecasting but there is no provision on how the model is updated incrementally. The method mentioned above, and other existing data-driven methods ignore the fact that they use future samples to calculate the loss function incrementally. Wu *et al.* [6] proposed a valid incremental learning adaptation framework with *LightGBM* where the model is only updated if there is a concept drift (prediction accuracy drops to a certain threshold), then a corrective factor is used to update the model incrementally. Even though this method does not consider future samples, it requires plenty of waiting time for enough data stream, before it can update its learning algorithm. In this study, we propose a fully incremental learning method motivated by piecewise linear curve fitting [7] and pseudo targeting [8]. Owoeye used pseudo targets to forecast the duration of stopover of migratory birds [9]. Zhao *et al.* used pseudo targets for trajectory forecasting of autonomous vehicles [10]. Currently, there are no time series multistep forecasting models in battery prognosis that use pseudo targets.

The key contribution of this paper is as follows: (i) model criteria to avoid data leakage from the future when building incremental multistep forecasting approach; (ii) unlike the existing FSNet, the proposed iFSNet uses pseudo target to train the model which is beneficial when it is deployed real-time; (iii) the proposed iFSNet applied on real-time battery capacity degradation datasets in an online mode; and (iv) incorporation of an adaptive learning rate for incremental learning.

The succeeding sections of this paper are organized in this order: Section II briefly reviews the current state-of-the-art methods in battery prognosis and online learning; Section III discusses some preliminaries on multistep forecasting and incremental multistep forecasting; Section IV introduces our proposed framework with some discussions on the limitations of the base architecture; Section V discusses our simulation setup to test our proposed framework on different battery degradation datasets; Section VI presents the results of the simulations and discusses their implications; lastly, Section VII summarizes our study and enumerates potential future work.

## II. STATE-OF-THE-ART

### A. Battery SoH Forecasting

Battery State-of-Health (SoH) is the ratio of battery capacity when fully charged, $Q_{full}$, and the battery capacity when it was brand new, $Q_{nominal}$, given by [11]:

$$SoH = \frac{Q_{full}}{Q_{nominal}} \times 100\% \quad (1)$$

The above equation is a common indicator to estimate the age of the battery, whether it is as good as new or if it is nearing its end-of-life.

Existing methods for battery SoH multistep forecasting were developed, but there is no emphasis on univariate time series models. Liu *et al.* [12] proposed a generalizable online battery degradation forecasting model which uses trend and cycle decomposition for forecasting and achieved a 2.8% average error evaluated on 35 batteries of MIT battery dataset [12]. However, the model is a multivariate time series approach since it also needs voltage data apart from the capacity data in order to forecast. Eaty *et al.* developed a continual learning approach to forecast battery SoH on 8 Li-ion batteries from NASA dataset and achieved 0.022 mean-squared error, equivalent to 0.1483 RMSE [13]. However, the model performs a multivariate time series task since it needs voltage, current, and temperature data as inputs apart from the capacity data in order to forecast. Our proposed framework is a univariate time series forecasting model which has less dependency on multiple uncertainty over measurements [14] in contrast to multivariate models.

### B. Online Learning

In traditional offline batch learning, the model is trained on the entire data before deployment. Therefore, post-deployment, an offline learning model must be retrained occasionally with varying data distribution. On the other hand, online learning continually train the model post-deployment [4]. The two schemes for online learning are incremental and mini batch schemes. Incremental learning retrains the deployed model on a single-pass mode (sample-by-sample) [15] as shown in Fig. 1. An alternative to incremental learning is mini batch learning [16], where the model must wait for a certain number of new streaming data, equivalent to the mini batch size, to arrive before it gets retrained. The downsides of this approach are the delay in retraining and limited forecast length which are both determined by the chosen mini batch size. This approach is beyond the scope of this study, but it is a reliable alternative to incremental learning in certain situations.

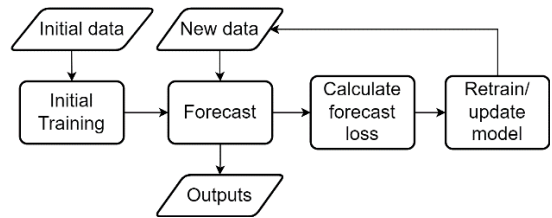

Figure 1. Flowchart of incremental learning.

Most existing online ML methods were applied to tasks with static data such as image detection from aerial image data [17] and defect classification for semiconductor failure analysis data [18]. However, these methods are for static datasets such as image segmentation and tabular data classification. There is not much emphasis on online learning for time series data. Hence, we focus on developing an online model for a dynamic dataset such as battery degradation data.

## III. PRELIMINARIES

### A. Multistep Ahead Forecasting

Univariate multistep ahead time series forecasting problem is given by $N$ input sequence ($\boldsymbol{x}$) that has to forecast $H$ output sequence ($\boldsymbol{y}$).

$$\boldsymbol{x}: x_{i-N}, x_{i-N-1}, \dots, x_{i-1} \quad (2)$$

$$\boldsymbol{y}: y_i, y_{i+1}, \dots, y_{i+H-1}, y_{i+H} \quad (3)$$

When $H$=1, the task becomes a one-step ahead time series forecasting. Applying incremental learning to it is straightforward. However, the impact and usability of one-step ahead forecasting is low compared to multistep ahead forecasting. In the context of prognosis and health management, one-step ahead forecasting might be too late to alert the users that the battery needs to be replaced. Even if the prediction is accurate, there may not be enough time before failure happens. Thus, multistep forecasting is preferable.

There are four main strategies for multistep ahead forecasting. The *direct strategy* involves developing multiple models, one model for each step-ahead forecast [19]. This strategy is not feasible for incremental learning as it requires $H$ models for an $H$-step forecasting task. The *recursive strategy* involves using a one-step ahead forecasting model multiple times by treating the recent prediction as the next sample until the forecast length is achieved [19]. This strategy only requires a single model, but its accuracy suffers from recursive error propagation. The *direct-recursive strategy* is a combination of the first two methods and is likewise not feasible for incremental learning [20]. The *multiple output strategy* [21] is the modern way of performing multistep ahead forecasting because most algorithms have a built-in method to perform such tasks. This strategy only requires a single model without error propagation unlike recursive strategy. For such reasons, only the multiple output strategy is explored in this study.

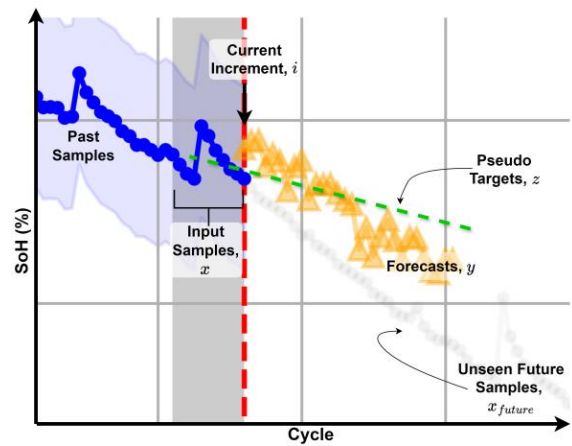

Figure 2. Multistep forecasting illustration for past samples, future samples, and pseudo targets. Blue shaded area indicate 5% error.

*B. Incremental Multistep Forecasting*

Incremental learning for one-step ahead forecasting is a straightforward task. Similarly, multistep ahead forecasting for offline learning is also a straightforward task. However, the combination of both incremental learning and multistep ahead forecasting is a difficult problem mainly because at a given instant, only one sample is obtained so that the model can only be corrected according to the most recent forecast $y_i$. The rest of the forecast $y_{i+1}, \dots, y_{i+H}$ is yet to be corrected as shown in Fig. 2. There are three approaches for updating the model in incremental multistep forecasting:

- *Invalid Approach:* Use the recent and **future samples** to update the model and calculate the forecast accuracy. This approach must be avoided as we do not have future samples.
- *Approach 1:* Use the recent and **past samples** to update the model and calculate the forecast accuracy. This requires waiting a considerable amount of time as shown in Fig. 3a.
- *Approach 2:* Our proposed method is to use the recent and **future pseudo targets** to update the model, and only use the recent sample to calculate the forecast accuracy as shown in Fig. 3b.

## IV. PROPOSED FRAMEWORK

Our proposed framework is an incremental implementation of FSNet which uses future pseudo targets to update the model at each increment.

*A. FSNet Limitations*

Fast and Slow learning Network (FSNet) was introduced by Pham *et al.* [22] for online time series forecasting. It uses Temporal Convolutional Network as base architecture and has adaptive structure fast learning mechanism and the associative memory slow learning mechanism. FSNet's concept for incremental multistep forecasting is correct but the implementation falls into the invalid case. It has a fixed data split of 25% training and 75% testing for a given data. Initially, it assumes 25% of data is collected and trained before moving to sample-by-sample training. It uses a standard scaler that fits the training data and transforms both training and test data which assumes that the mean and standard deviation of the test data is close to that of training data. This assumption is not valid for battery degradation data where the trajectory is a downward trend, not a periodic time series.

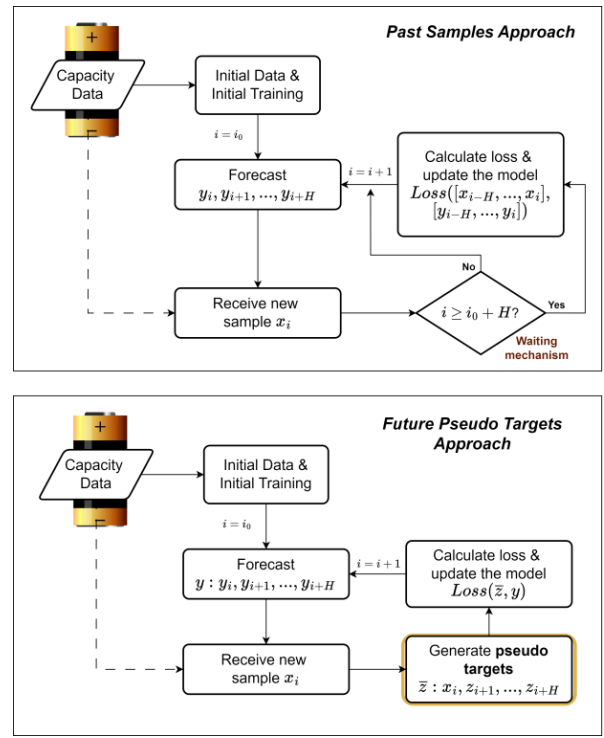

Figure 3. Two valid approach for incremental multistep forecasting (a) using past samples and (b) using future pseudo targets – no need to wait as pseudo target is passed to the model.

*B. Incremental FSNet*

Our proposed incremental implementation of FSNet involves using pseudo targets as a future reference to calculate the loss function from the forecasted values at each increment. Several pseudo targets $z$ can be chosen such as from moving average, Fourier-decomposed cycles, etc. In this study, we only considered a simple linear regressor for the best fit line of the input sequence since higher order polynomials worsen the model prediction.

$$\boldsymbol{x}_{input} = m\boldsymbol{t}_{input} + b \qquad (4)$$

Where $m$ is the slope and $b$ is the intercept. Since the slope cannot be increasing in a battery degradation trajectory, we define $m'$ as,

$$m' = \begin{cases} m, m < 0 \\ 0, m \geq 0 \end{cases} \qquad (5)$$

Once we calculate the slope $m'$ and intercept $b$, we extrapolate the next $H$ values on time $\boldsymbol{t}$ to generate the pseudo targets $z$ given by,

$$\boldsymbol{z} = m'\boldsymbol{t}_{forecast} + b. \qquad (6)$$

The forecast loss, *MSELoss*, for each increment can be calculated as,

$$MSELoss = \frac{1}{H}(x_i - y_i)^2 + \frac{1}{H}\sum_{j=i+1}^{i+H}(z_j - y_j)^2 \quad (7)$$

where the first pseudo target was replaced by the actual new sample $x_i$ received from the streaming data. After the streaming data was completed, a post analysis was done to evaluate the forecast performance of the models. We used the common metrics for battery SoH prediction, root mean-squared error (RMSE) and mean absolute error (MAE) given by the following equations:

$$RMSE = \sqrt{\frac{1}{n}\sum_{j=1}^{n}(x_j - y_j)^2} \quad (8)$$

$$MAE = \frac{1}{n}\sum_{j=1}^{n}|x_j - y_j| \quad (9)$$

We tried to further improve our proposed model by introducing a factor that reduces the learning rate during retraining. The $\gamma$ factor is the ratio of the error, $RMSE_{inc}$, of the forecast at current instance and the error, $RMSE_{pseudo}$, of the pseudo target at current instance, which is multiplied by the learning rate.

$$\gamma = \frac{RMSE_{inc}}{RMSE_{pseudo}} \quad (10)$$

$$\eta = \begin{cases} \eta_0, & \gamma \geq 1 \\ 0.1\eta_0\gamma, & \gamma < 1 \end{cases} \quad (11)$$

The $\gamma$ factor represents the confidence of the model on the pseudo targets of the current iteration. Thus, $\gamma \in (0, +\infty)$ reduces the learning rate $\eta \in (0, \eta_0)$ when the model has low confidence, $\gamma < 1$, in the pseudo targets. Otherwise, the model uses the initial learning rate, $\eta_0 = 10^{-5}$ which prevents the learning rate from blowing up to $+\infty$.

*C. Incremental Multistep Forecasting Criteria*

Key considerations for multistep ahead forecasting models with incremental learning mechanisms:
- ✓ The model does not use future samples to calculate the loss function of the entire forecast for retraining/updating at each increment.
- ✓ The model does not use future samples to calculate the model performance of the entire forecast at each increment.
- ✓ The model does not use the streaming data more than once to train the online model [15]. This is permissible with the caution of overfitting the training data.

V. TEST SIMULATION SETUP

We tested our proposed framework against models that represent the traditional and modern forecasting methods on forecast accuracy and iteration run time complexity. We tested each model using several experimental data of different characteristics. We also evaluated the algorithm's sensitivity to

TABLE I. DESCRIPTION OF THE BATTERY DATASETS.

| Data | Nominal Capacity | Battery Lifetime | Smoothness | Knee point |
|---|---|---|---|---|
| MIT B1C21 | 1.1 Ah | Short (557 cycles) | Smooth | Yes |
| NASA B0005 | 2.0 Ah | Short (168 cycles) | Irregular[a] | No |
| MIT B1C3 | 1.1 Ah | Long (1224 cycles) | Smooth | Yes |
| CALCE CX2-36 | 1.35 Ah | Long (1887 cycles) | Irregular[a] | No |

a. Irregular non-smooth trajectory with capacity regeneration spikes.

the size of the input sequence (lookback window) and to the forecast length (forecast horizon). The simulations were run on a standard workstation of 12[th] Gen Intel(R) Core (TM) i9-12900H CPU and NVIDIA GeForce RTX 3070 Ti Laptop GPU.

*A. Data*

We used experimental Lithium-ion battery degradation data from publicly available datasets from the National Aeronautics and Space Administration Prognostics Center of Excellence (NASA PCoE) [23], Massachusetts Institute of Technology (MIT) by Severson *et al.* [24], and Center for Advanced Life Cycle Engineering (CALCE) [25]. We tested our model on four battery degradation data which are carefully selected to represent each permutations of short or long lifetime, and smooth or with capacity regeneration spikes [26] summarized in Table I. The MIT B1C21 battery data has a short lifetime of 557 charge-discharge cycles and with smooth trajectory, NASA B0005 battery data has a short lifetime of 168 charge-discharge cycles with capacity regeneration spikes, MIT B1C3 battery data has a long lifetime of 1224 charge-discharge cycles with smooth trajectory, and CALCE CX2-36 battery data has a long lifetime of 1887 charge-discharge cycles and a non-smooth trajectory with capacity regeneration spikes. Additionally, we used the NASA B0006 battery to pretrain the learning models that will be tested on NASA B0005 because B0005 only has a few cycles. NASA B0006 was chosen because it has the same battery chemistry, battery geometry, and nominal capacity as NASA B0005. The rest of the battery datasets have enough cycles for warm-up training and, thus, there is no need for pretraining on separate battery dataset.

*B. Test Models*

We compared our proposed model against several existing models, each representing a particular case. Our hypotheses are as follows: first, our model performs better than a non-learning model; second, our model performs better than a model which uses past samples to retrain incrementally; third, our model performs better than a model that has no framework to mitigate concept drift and catastrophic forgetting; lastly, we also tested if the iFSNet-$\gamma$, by incorporating equation 11, improves the performance of iFSNet. We set the input length *N*=10 and the forecast length *H*=30. The choice of the forecast length is approximately in between the two test values, $H = \{24, 48\}$, that FSNet used in their paper [22] for multistep forecasting.

TABLE II. TEST MODELS REPRESENT MODELS THAT ARE LEARNING/NON-LEARNING, WITH/WITHOUT CONCEPT DRIFT MECHANISM, AND DIFFERENT INCREMENTAL MULTISTEP FORECASTING (IMF) APPROACH.

| Model | Learning | w/ Concept Drift Mechanism | IMF Approach |
|---|---|---|---|
| ARIMA | No | No | None |
| RNN | Yes | No | Pseudo targets |
| FSNet-corrected | Yes | Yes | Past samples |
| iFSNet/iFSNet-$\gamma$ | Yes | Yes | Pseudo targets |

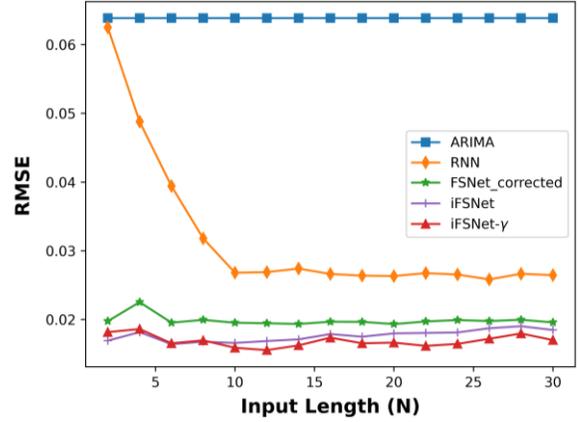

The forecast length $H = 30$, could also represent one month ahead forecast if the battery is cycled once a day.

**ARIMA**. We used Autoregressive Integrated Moving Average (ARIMA) [27] to represent forecasting models that have no learning mechanism, that is, the model fits the input samples and then forecasts without correcting the mechanism on whether the forecasts are accurate or not. The choice for ARIMA model's seasonality, $m$, is arbitrary since it is predetermined a priori. This is one drawback of using ARIMA in an online learning task. In this study, we used $m$=7 which is an arbitrary number lower than the input length $N$=10. The choice of $m$ could represent a weekly seasonality if the battery is cycled once a day.

**FSNet-corrected**. We incorporated a waiting mechanism for FSNet to correct the implementation of incremental learning. Instead of immediate retraining after forecasting, FSNet-corrected waits until there are enough new samples to calculate the loss from the previous. In this way, FSNet-corrected still forecasts at every increment but the model update is delayed once. For the succeeding increments, the model will both forecast $y$ and update using past samples. Since the model is trained with the past samples repeatedly, it is prone to overfitting.

TABLE III. CONFIGURATION & HYPERPARAMETERS.

| Configuration | Value |
|---|---|
| **ARIMA** | |
| Seasonal | True |
| Parameter search | *AutoARIMA* |
| m | 7 |
| **FSNet-corrected, iFSNet, iFSNet-$\gamma$** | |
| Optimizer | Adam |
| Warm-up Learning Rate | 0.001 |
| Update Learning Rate ($\eta_0$) | 1e-5 |
| Train epochs | 8 |
| Loss Function | *MSELoss* |
| **RNN** | |
| Optimizer | SGD |
| Learning Rate (LR) | 0.1 |
| LR Decay Rate | 0.9 |
| LR Decay Rate Steps | 1000 |
| Momentum | 0.9 |
| Loss Function | *MSELoss* |
| LR Scheduler | Exponential |

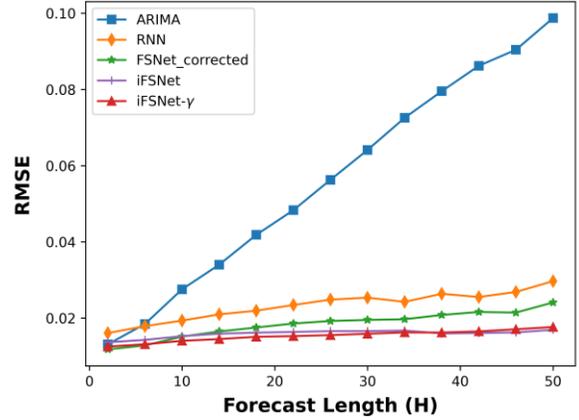

Figure 4. RMSE of each model on **CALCE CX2-36 dataset** for (top) various input length $N$ with forecast length $H$=30, and (bottom) various forecast length $H$ with input length $N$=10.

**RNN**. We used Recurrent Neural Network (RNN) [28] to represent forecasting models that have neural network structure but do not handle concept drift and catastrophic forgetting. The incremental updating mechanism of this model is also based on the pseudo targets that are similar to iFSNet but the only difference from iFSNet is the absence of concept drift and catastrophic forgetting. The RNN architecture includes RNN layers of 50 units, a dropout layer of 0.2 rate, fully connected layer, and a Sigmoid activation function. The hyperparameters are summarized in Table III.

## VI. RESULTS AND DISCUSSION

### A. Sensitivity to Input Length and Forecast Length

From Fig. 4 (top image), we can observe that FSNet-corrected, iFSNet, and iFSNet-$\gamma$ achieve their optimal performance when the input length $N \geq 6$. Likewise, RNN performs worse when the input length is too short and then converges to better RMSE when the input length $N \geq 10$. ARIMA's performance is not dependent on input length because it does not use a look back window, instead, it trains on the entire past samples incrementally with its *update* function.

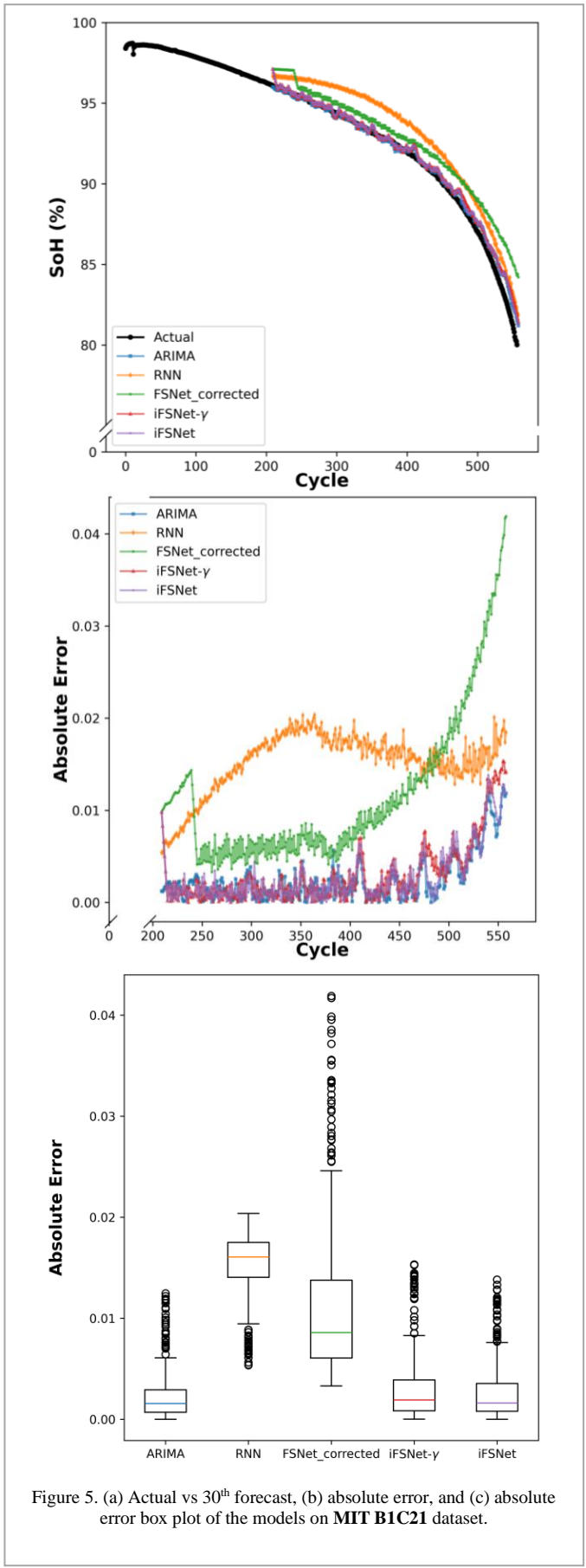

Figure 5. (a) Actual vs 30$^{th}$ forecast, (b) absolute error, and (c) absolute error box plot of the models on **MIT B1C21** dataset.

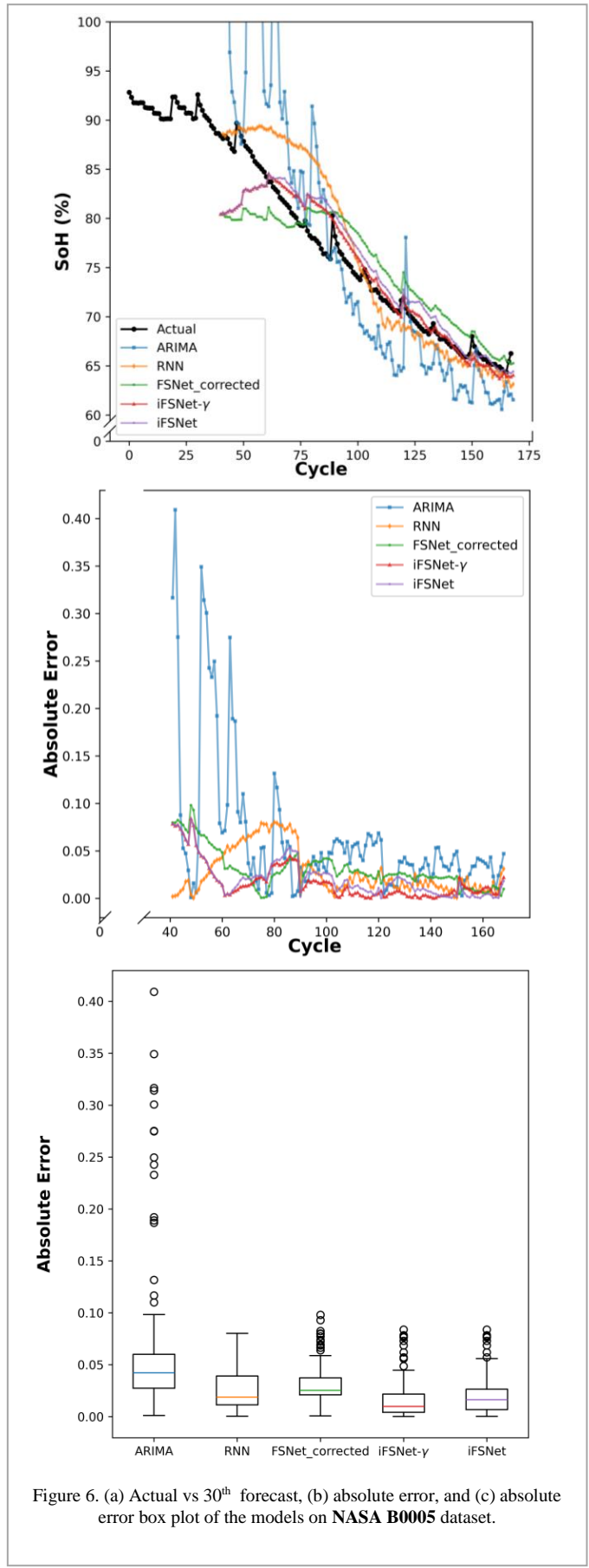

Figure 6. (a) Actual vs 30$^{th}$ forecast, (b) absolute error, and (c) absolute error box plot of the models on **NASA B0005** dataset.

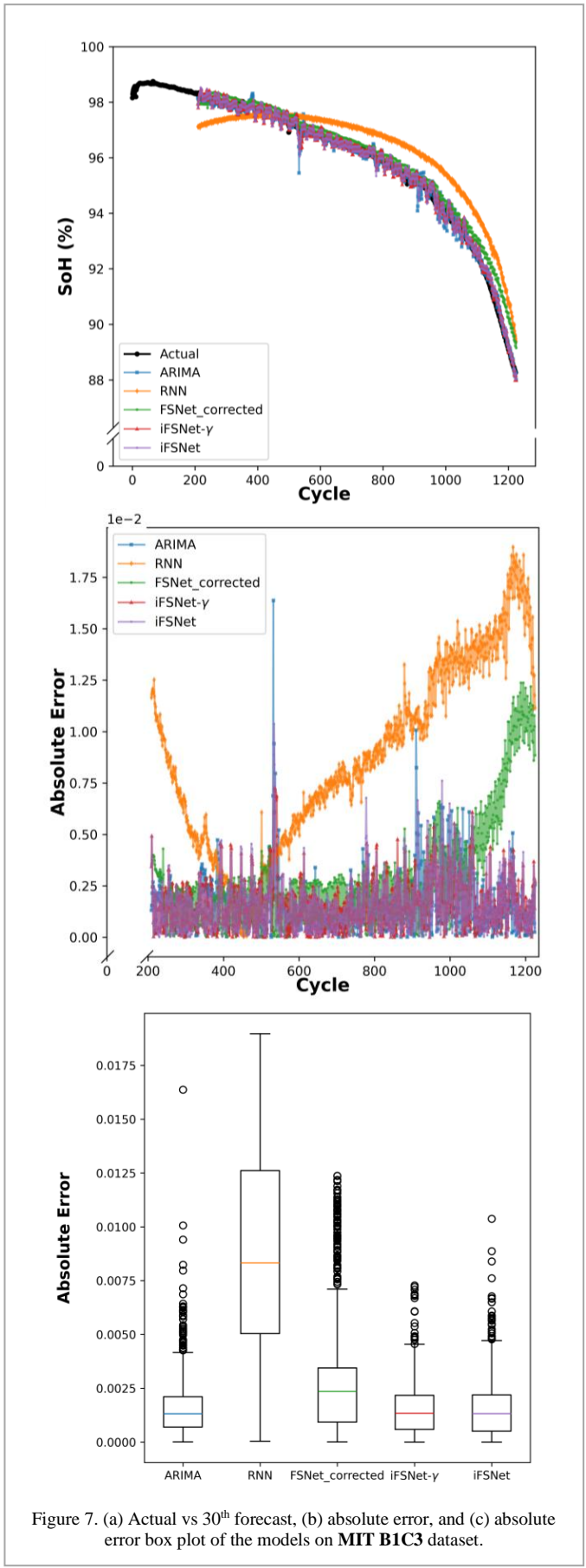

Figure 7. (a) Actual vs 30[th] forecast, (b) absolute error, and (c) absolute error box plot of the models on **MIT B1C3** dataset.

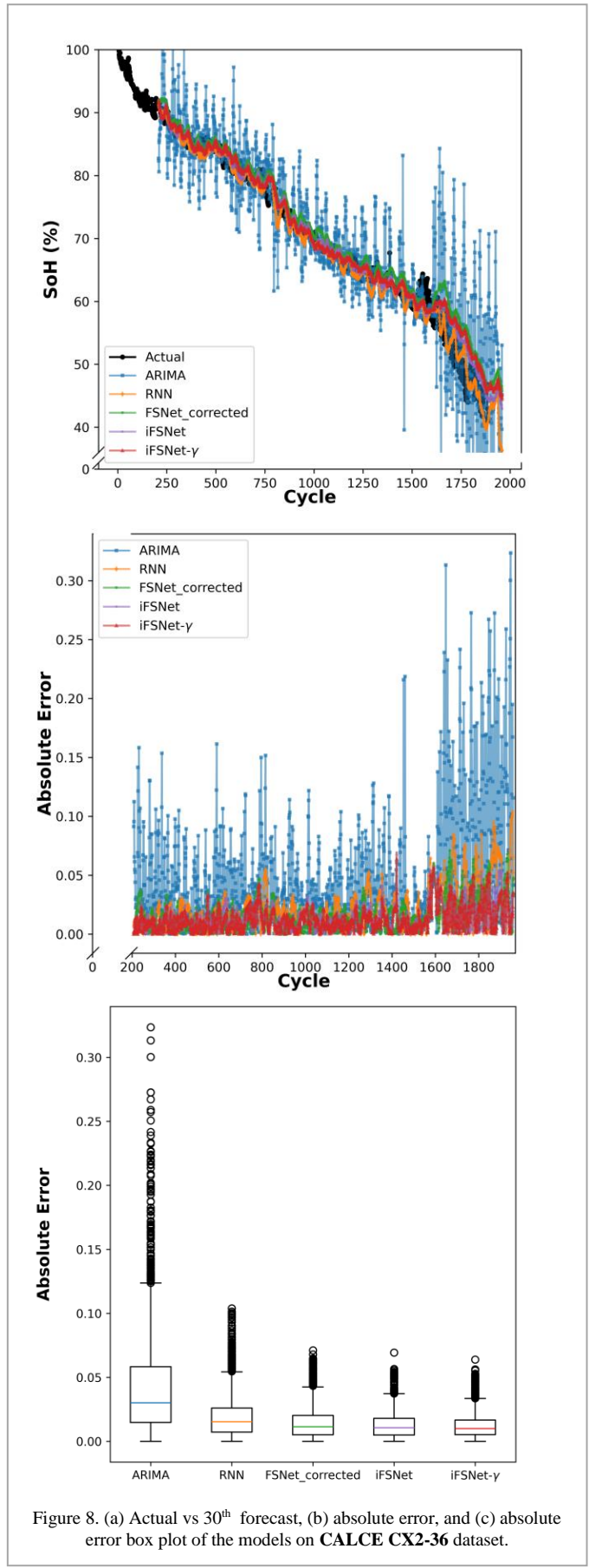

Figure 8. (a) Actual vs 30[th] forecast, (b) absolute error, and (c) absolute error box plot of the models on **CALCE CX2-36** dataset.

TABLE IV. FORECASTING ACCURACY AND RUN TIME COMPLEXITY COMPARISON BETWEEN IFSNET AND OTHER THE MODELS.
INPUT SEQUENCE LENGTH N=10 AND FORECAST LENGTH H=30.

| Data | ARIMA | | | RNN | | | FSNet-corrected | | | iFSNet | | | iFSNet-$\gamma$ | | |
|---|---|---|---|---|---|---|---|---|---|---|---|---|---|---|---|
| | RMSE | MAE % | Time s/it | RMSE | MAE % | Time s/it | RMSE | MAE % | Time s/it | RMSE | MAE % | Time s/it | RMSE | MAE % | Time s/it |
| *MIT (B1C21)* | **0.00354** | **0.241** | 6.06 | 0.01560 | 1.525 | **0.0028** | 0.01433 | 0.173 | 0.0661 | 0.00401 | 0.275 | 0.0449 | 0.00454 | 0.304 | 0.0459 |
| *NASA (B0005)* | 0.10008 | 6.484 | 1.78 | 0.03681 | 2.820 | **0.0036** | 0.03654 | 0.042 | 0.0568 | 0.02884 | 0.105 | 0.0429 | **0.02642** | 0.766 | 0.0423 |
| *MIT (B1C3)* | 0.00207 | 0.158 | 2.96 | 0.00976 | 0.864 | **0.0026** | 0.00393 | 0.288 | 0.0899 | 0.00207 | 0.157 | 0.0437 | **0.00197** | **0.154** | 0.0410 |
| *CALCE (CX2-36)* | 0.06385 | 4.468 | 1.13 | 0.02677 | 1.976 | **0.0024** | 0.01952 | 0.470 | 0.1113 | 0.01658 | 0.288 | 0.0418 | **0.01588** | **0.234** | 0.0420 |

From Fig. 4 (bottom image), we can observe that all the models have increasing RMSE with increasing forecast length. However, ARIMA has the highest rate of increase in error as the forecast length increases. ARIMA has comparable performance to the learning models when the forecast length $H < 10$. The models, RNN, iFSNet, and iFSNet$\gamma$, which used pseudo targets are expected to have increasing RMSE as the forecasting length increases since the extrapolated linear regression is expected to be increasingly off the true values as the best fit line is extended increasingly.

*B. iFSNet Compared to other Models*

Results in Figures 5-8 show the (a) actual vs the $H^{th}$ forecast, where all the experiments $H$ is set as 30, of each model in the four battery datasets, (b) absolute error plot and (c) box plots for absolute error. The absolute error is given by MAE in equation 9 without taking the mean, i.e. take the absolute value of the difference between the prediction and the true value. Only the $30^{th}$ forecast is shown to avoid overplotting, since plotting all the forecasts of all the models at each increment would make the graphs incomprehensible.

ARIMA forecasts well on smooth battery trajectories (blue line in Fig. 5 and Fig. 7) achieving 0.00207-0.00354 RMSE on MIT datasets (Table IV) while it struggles on irregular battery trajectories (blue line in Fig. 6 and Fig. 8) achieving 0.1 RMSE and 6.484% MAE on NASA B0005 and 0.06385 RMSE and 4.468% MAE on CALCE CX2-36. The absolute error of its forecasts ranges from 0.1-0.4 on non-smooth trajectories. ARIMA remains a good baseline for time series forecasting models. It unexpectedly outperformed all the models in the MIT B1C21 dataset with 0.001 RMSE and 0.06% MAE lower than iFSNet. This is because ARIMA assumes stationarity in the time series while it is forecasting, and MIT datasets are stationary, while the datasets with capacity regeneration spikes are non-stationary. ARIMA has the slowest run time of 1.13-6.06 seconds per iteration (1.09-6.02 seconds slower than iFSNet) among the models due to the automated parameter search of *AutoARIMA* for ARIMA's primary parameters (p,d,q) and three seasonal parameters (P,D,Q) which vary per iteration.

RNN has the fastest run times of 0.0024-0.0028 seconds per iteration due to the simplicity of its design. However, since it has no mechanism against concept drift and catastrophic forgetting, it performs worse than the other models. On average, its performance is closest to but slightly worse than FSNet-corrected.

FSNet-corrected is still a reliable model achieving 0.036 RMSE on the NASA dataset and 0.014 RMSE on the CALCE dataset. However, it struggles on datasets with visible knee points. It overfits the past samples by training on them multiple times resulting in worse forecasts. The overfitting will even be worse for periodic time series datasets that go on indefinitely.

Considering all the cases, iFSNet and iFSNet-$\gamma$ are the most robust and consistent in terms of forecast accuracy and run time complexity. The iFSNet-$\gamma$ outperformed all the models in three datasets achieving 0.026 RMSE on the NASA dataset, 0.0019 RMSE on the MIT B1C3 dataset, and 0.015 RMSE on the CALCE dataset. The proposed iFSNet performs well since it captures the short-term time series dependencies through its TCN base model and the long-term dependencies through its associative memory mechanism which handles long-term recurring patterns [22]. The use of pseudo targets also improve the model since the pseudo targets represent the hidden future samples as multiple linear regression with unknown break points [7]. Both iFSNet and iFSNet-$\gamma$ have a consistent run time complexity of approximately 0.04 seconds per iteration which is efficient enough for online implementation. This is due to the adaptive structure mechanism of the FSNet where each layer begins with shallow architecture and adapts independently [22]. The improved version, iFSNet-$\gamma$, has all the features of iFSNet with additional adaptive mechanism that reduces the learning rate when the pseudo target has low confidence. This prevents the model from getting worse due to inaccurate pseudo targets.

*C. Discussion and Summary*

We can observe from the results that FSNet performs worse than expected when the incremental implementation was corrected with waiting mechanisms. Similar mistakes are common in the traditional offline batch machine learning referred to as *no-time-machine* data leakage by Kaufman *et al.* [29] and *leaks from the future* by Nisbet *et al.* [30]. However, such oversights are subtle in the context of online learning where the leakages occur during the training process while optimizing the weights of the model.

This study uses a univariate time series model where the only input needed is the capacity measurement converted to SoH. A univariate model has an advantage over a multivariate model in terms of robustness to real-world noisy data since it only depends on one measurement in contrast to the multiple input feature dependence of multivariate models. For example, a multivariate model that requires input measurements of voltage, current, temperature, and capacity will suffer if voltage measurements are inaccurate, while the univariate model is not affected. Moreover, if the capacity measurement is inaccurate then both models will perform poorly. In both cases, the univariate model is more robust than the multivariate model in real-world situations of imperfect data measurements.

Our model was tested on battery capacity degradation data where the trajectory is a downward trend and halts at a certain point, not the typical periodic time series data. The forecast length is limited to the initial training data, e.g. if the desired forecast length is 30, then the initial training should have more than 30 samples. In circumstances where no data from a similar battery is available, initial training data can be sample-by-sample online learning approach. In such cases, the forecasts will have larger errors at the beginning and the online model might not learn unless the learning rate is increased leading to instability. Other schemes such as direct, recursive, mini batch, and their combinations are not explored in this study. In the context of deployment, there is no prior knowledge if the battery life will be short or long, thus, pretraining on another similar battery data – battery chemistry, battery geometry, and nominal capacity – if available, is a better option.

## VII. Conclusion

In this study, we investigated the common mistakes of implementing incremental multistep forecasting. We showed how existing methods have lower forecast accuracy than reported because they failed to incorporate a waiting mechanism to avoid using future samples. The approach presented in this work can be applied to any model, not limited to FSNet, for proper implementation of incremental learning on multistep ahead forecasting and proper evaluation of forecasting accuracy. Our proposed model benefits from the adaptive structure and associative memory mechanisms of FSNet and at the same time incrementally improves by using pseudo targets. Our proposed iFSNet-$\gamma$ achieved 0.00197 RMSE and 0.00154 MAE on battery datasets with smooth degradation trajectories while it achieved 0.01588 RMSE and 0.01234 MAE on battery datasets with irregular degradation trajectories with capacity regeneration spikes. For comparison, iFSNet had an RMSE that is 97.2% lower than [13] on the NASA battery dataset.

Future studies can be done to evaluate our proposed pseudo target approach for online learning on typical periodic time series datasets that continue indefinitely such as traffic, rainfall, and electrical energy consumption datasets.

## Acknowledgment

The first author acknowledges Singapore University of Technology and Design (SUTD) and the Ministry of Education, Singapore (MOE) for the Research Student Scholarship (RSS) and SUTD Ph.D. Fellowship for his doctoral studies from 2023-2026. The logistics of this research is also supported by the Research Surplus Grant No. RGSUR08 of Prof. Nagarajan Raghavan. The authors would also like to thank the Agency for Science, Technology, and Research (A*STAR) for providing computational resources for this study.

[14] measurement technology," *Journal of Power Sources*, vol. 561, p. 232749, Mar. 2023.

[15] C. Giraud-Carrier, "A Note on the Utility of Incremental Learning," *AI Communications*, 2004.

[16] Z. Yang, C. Wang, Z. Zhang, and J. Li, "Mini-batch algorithms with online step size," *Knowledge-Based Systems*, vol. 165, pp. 228–240, Feb. 2019.

[17] J. Senthilnath *et al.*, "BS-McL: Bilevel Segmentation Framework With Metacognitive Learning for Detection of the Power Lines in UAV Imagery," *IEEE Trans. Geosci. Remote Sensing*, vol. 60, pp. 1–12, 2022.

[18] B. Zhou, P. Jieming, M. Sivan, A. V.-Y. Thean, and J. Senthilnath, "Quantile Online Learning for Semiconductor Failure Analysis," in *ICASSP 2023 - 2023 IEEE International Conference on Acoustics, Speech and Signal Processing (ICASSP)*, Rhodes Island, Greece: IEEE, Jun. 2023, pp. 1–5.

[19] G. C. Tiao and R. S. Tsay, "Some advances in non-linear and adaptive modelling in time-series," *Journal of Forecasting*, vol. 13, no. 2, pp. 109–131, Mar. 1994.

[20] A. Sorjamaa and A. Lendasse, "Time Series Prediction using DirRec Strategy," 2006.

[21] R. B. Kline, *Beyond significance testing: Reforming data analysis methods in behavioral research*. Washington: American Psychological Association, 2004.

[22] Q. Pham, C. Liu, D. Sahoo, and S. C. H. Hoi, "Learning Fast and Slow for Online Time Series Forecasting," *International Conference on Learning Representations*, 2023.

[23] B. Saha and K. Goebel, "Battery dataset." NASA AMES prognostics data repository, 2007.

[24] K. A. Severson *et al.*, "Data-driven prediction of battery cycle life before capacity degradation," *Nat Energy*, vol. 4, no. 5, pp. 383–391, Mar. 2019.

[25] "Battery Data | Center for Advanced Life Cycle Engineering." Accessed: Jul. 25, 2024. [Online]. Available: https://calce.umd.edu/battery-data

[26] J. Wilhelm *et al.*, "Cycling capacity recovery effect: A coulombic efficiency and post-mortem study," *Journal of Power Sources*, vol. 365, pp. 327–338, Oct. 2017.

[27] G. Box and G. Jenkins, *Time Series Analysis: Forecasting and Control*. San Francisco, California: Holden-Day Inc., 1970.

[28] J. J. Hopfield, "Neural networks and physical systems with emergent collective computational abilities.," *Proc. Natl. Acad. Sci. U.S.A.*, vol. 79, no. 8, pp. 2554–2558, Apr. 1982.

[29] S. Kaufman, S. Rosset, C. Perlich, and O. Stitelman, "Leakage in data mining: Formulation, detection, and avoidance," *ACM Trans. Knowl. Discov. Data*, vol. 6, no. 4, pp. 1–21, Dec. 2012.

[30] R. Nisbet, G. Miner, and K. P. Yale, *Handbook of statistical analysis and data mining applications*, Second edition. London, United Kingdom: Academic Press, an imprint of Elsevier, 2018.